\documentclass[runningheads]{llncs}
\usepackage{graphicx}
\usepackage{amsmath}
\usepackage{adjustbox}
\usepackage{algorithm}
\usepackage{algpseudocode}
\usepackage{subcaption}
\usepackage{booktabs} 
\usepackage{longtable} 
\usepackage{comment}
\usepackage{xcolor}
\usepackage{mdframed}
\usepackage[utf8]{inputenc}
\usepackage{float} 
\usepackage{comment}

\usepackage{hyperref}

\begin{document}
\title{Unveiling Disparities in Maternity Care: A Topic Modelling Approach to Analysing Maternity Incident Investigation Reports}
\titlerunning{Unveiling Disparities in Maternity Care}
%

\author{Georgina Cosma\inst{1} \and
Mohit Kumar~Singh\inst{1} \and
Patrick Waterson\inst{2} \and Gyuchan Thomas Jun\inst{2}\and Jonathan Back \inst{3}}
\authorrunning{G. Cosma et al.}

\institute{Department of Computer Science, School of Science, Loughborough University, UK \and
School of Design and Creating Arts, Loughborough University, UK \and
Health Services Safety Investigations Body (HSSIB), United Kingdom
\email{g.cosma@lboro.ac.uk}\\
}

\maketitle             

\begin{abstract}
This study applies Natural Language Processing techniques, including Latent Dirichlet Allocation, to analyse anonymised maternity incident investigation reports from the Healthcare Safety Investigation Branch. The reports underwent preprocessing, annotation using the Safety Intelligence Research taxonomy, and topic modelling to uncover prevalent topics and detect differences in maternity care across ethnic groups. A combination of offline and online methods was utilised to ensure data protection whilst enabling advanced analysis, with offline processing for sensitive data and online processing for non-sensitive data using the `Claude 3 Opus' language model. Interactive topic analysis and semantic network visualisation were employed to extract and display thematic topics and visualise semantic relationships among keywords. The analysis revealed disparities in care among different ethnic groups, with distinct focus areas for the Black, Asian, and White British ethnic groups. The study demonstrates the effectiveness of topic modelling and NLP techniques in analysing maternity incident investigation reports and highlighting disparities in care. The findings emphasise the crucial role of advanced data analysis in improving maternity care quality and equity.

\end{abstract}

\section{Introduction}
Maternity hospital deaths have been a significant concern in the United Kingdom, and the prevention of future deaths (PFD) reports have played a crucial role in addressing this issue. The Healthcare Safety Investigation Branch (HSIB) was an independent organisation established to investigate patient safety incidents within the NHS in England. Established in 2017 and funded by the Department of Health and Social Care, HSIB aimed to improve patient safety through its investigations. In October 2023, HSIB transformed into two organisations: the Maternity and Newborn Safety Investigations (MNSI), which is hosted by the Care Quality Commission, and the Health Services Safety Investigations Body (HSSIB), which is an independent statutory body. HSIB's maternity investigation programme focused on identifying systemic factors contributing to adverse events in maternity care. The programme investigated qualifying incidents referred by NHS trusts, such as intrapartum stillbirths, early neonatal deaths, and severe brain injuries in term babies. The HSIB Maternity Investigation Programme Year in Review 2022/23 \cite{HSIB2023} provides an overview of the programme's activities and findings from 1 April 2022 to 31 March 2023. During this period, HSIB received 1\,070 referrals for maternity incidents, of which 671 progressed to full investigations. The most common type of incident investigated was term babies who were cooled or diagnosed with brain injuries (52\%), followed by intrapartum stillbirths (26\%), early neonatal deaths (13\%), and maternal deaths (9\%). The review highlights a reduction in referrals for severe brain injury in babies over the past few years, with a 13\% decrease in 2022/23 compared to the previous year and a 19\% decline compared to 2019/20. The investigations also identified recurring themes, such as issues with clinical assessment, guidance adherence, foetal monitoring, escalation protocols, and communication among healthcare professionals. One of the most high-profile cases that led to a PFD report was the death of baby Harry Richford at East Kent Hospitals University NHS Foundation Trust in 2017 \cite{RichfordInquest2020}. The coroner's report outlines numerous failures and concerns that include the inadequate assessment and supervision of a locum registrar, confusion among staff regarding guidelines and when to escalate care, insufficient neonatal resuscitation training, poor record-keeping, and failure to properly report the death. The coroner made 19 recommendations for the Trust and national bodies to address these problems, such as reviewing locum recruitment, clarifying policies, improving training, and auditing practices, to prevent future deaths. \\
The Ockenden Review, an independent inquiry into maternity services at the Shrewsbury and Telford Hospital NHS Trust, was initiated due to concerns raised by the families of babies who died or were harmed in the maternity units of the Trust \cite{ockenden2020final}. The Ockenden Report scrutinises maternity services at Shrewsbury and Telford Hospital NHS Trust from 2000 to 2019, covering 1\,486 families and 1\,592 clinical incidents. The report highlights a culture of poor communication with affected families and insufficient adherence to clinical guidelines, underlining an urgent need for improved leadership, transparency, and a comprehensive maternity workforce plan to prevent future failings. \\
The Royal College of Obstetricians and Gynaecologists (RCOG) has also taken steps to address the issues highlighted in PFD reports. The ``Each Baby Counts: 2020 Final Progress Report'' \cite{RCOG2020EachBabyCounts} by the Royal College of Obstetricians and Gynaecologists (RCOG) presents a critical evaluation of maternity care outcomes in the UK, focusing on 2018 data involving 1\,145 cases of stillbirths, neonatal deaths, and severe brain injuries during term labour. A pivotal finding was that in 74\% of cases, different care might have led to different outcomes, suggesting systemic deficiencies in clinical governance, risk assessment, and neonatal care. The report underscores the importance of multidisciplinary reviews and enhanced parental involvement, with 70\% of parents invited to contribute to the review process. Despite the challenges in achieving its ambitious goal to halve these tragic incidents by 2020, the initiative significantly raised maternity safety awareness and set a foundation for future improvements. The Each Baby Counts programme, through its comprehensive analysis and recommendations, leaves a lasting legacy, advocating for sustained efforts to address disparities and improve maternity care services across the UK.\\
This paper utilises a range of Natural Language Processing (NLP) techniques to analyse 188 maternity incident reports provided by HSIB to identify disparities in care across different ethnic groups, underscoring the critical need for reforms to enhance maternity care quality and equity.

\section{Methodology} \label{metho}
\subsection{Dataset of maternity incident investigation reports} 
The Healthcare Safety Investigation Branch (HSIB) provided a random set of 188 anonymised investigation reports describing adverse maternity incidents. The reports were written between 2019 to 2022. The number of reports for each year is as follows: 4 reports in 2019, 115 reports in 2020, 42 reports in 2021, and 27 reports in 2022. Ethnicity was provided for 76 reports. The reports were processed, and the output was a text file (in CSV format) containing File IDs, Sentence IDs, sentence, and concepts from the Safety Intelligence ResearCH (SIRch) taxonomy. During the processing phase, the files underwent cleaning, involving the selection of specific sentences from each report. The sentence selection process focused on identifying sentences that contained negative connotations, references to physical characteristics, and mentions of medication names related to dispensing medications. The selected sentences were annotated using the SIRch taxonomy and compiled into a CSV file, which is utilised for clustering and summarisation tasks. This concept annotation task was carried out using our I-SIRch tool, and the process is described in \cite{Singh2023I-SIRch}. Table \ref{Labels} shows the list of Concepts with the number of sentences per concept shown in brackets. 
\begin{table}[]
\centering
\caption{ID, concept and number of sentences per concept. Total sentences: 3760.}
\label{Labels}
\begin{tabular}{cp{2in}cp{2in}}
\toprule
\textbf{ID} & \textbf{Concept} & \textbf{Index} & \textbf{Label} \\
\midrule
1 & Acuity (54) & 2 & Antenatal (69) \\
3 & Assessment, investigation, testing, screening (381) & 4 & COVID (142) \\
5 & Care Planning (132) & 6 & Communication factor (477) \\
7 & Decision error (169) & 8 & Dispensing, administering (62) \\
9 & Documentation (168) & 10 & Escalation/referral factor (158) \\
11 & Guidance factor (42) & 12 & Interpretation (197) \\
13 & Language barrier (30) & 14 & Local guidance (88) \\
15 & Monitoring (118) & 16 & National and local guidance (80) \\
17 & National guidance (169) & 18 & Obstetric review (147) \\
19 & Physical characteristics (320) & 20 & Physical layout and Environment (35) \\
21 & Psychological characteristics (54) & 22 & Risk assessment (94) \\
23 & Situation awareness (77) & 24 & Slip or lapse (188) \\
25 & Teamworking (530) & 26 & Technologies and Tools-issues (112) \\
27 & Training and education (40) & & \\
\bottomrule
\end{tabular}
\end{table}

\subsection{Topic modelling and semantic network visualisation methods}
\subsubsection{Textual data analysis and topic modeling.} The methods proposed in this study focus on extracting and analysing topics from a preprocessed dataset using Natural Language Processing (NLP) techniques. The primary goal is to identify unique topics within the dataset based on different combinations of concepts and ethnic groups. The process is provided in Algorithm (\ref{alg1}). All methods discussed in Section \ref{metho} are offline. \\ \textit{Topic Extraction:} To extract topics from the dataset, the \texttt{extract\_topics\_lda} function is defined. This function takes the DataFrame, a specific concept, ethnicity, and the desired number of topics as input parameters. It subsets the dataset based on the provided concept and ethnicity values, allowing for the extraction of topics specific to particular combinations. \textit{TF-IDF Vectorisation:} The text data in the `Sentence' column of the subset is transformed into a numerical representation using the Term Frequency-Inverse Document Frequency (TF-IDF) vectorisation technique. The resulting TF-IDF matrix is then normalised using the L2 norm to eliminate discrepancies caused by document size differences in the TF-IDF matrix. \textit{Latent Dirichlet Allocation (LDA):} The LDA model is applied to the normalised TF-IDF matrix, specifying the desired number of topics. LDA is a probabilistic topic modelling algorithm that discovers latent topics within a collection of documents. It assigns each sentence to a mixture of topics and represents each topic as a distribution over words. \\
\textit{Top Words Retrieval:} After applying the LDA model, the top words associated with each topic are retrieved. These top words provide a summarised representation of the main topics discussed within each topic. \textit{Duplicate Topic Removal:} To remove duplicate topics the \texttt{remove\_duplicates} function is defined. It creates a new list containing only the unique topics from the input list, discarding any duplicates. \textit{Topic Storage:} The extracted topics, along with their corresponding concept, ethnicity values, and sentence count are stored in a dictionary. \textit{DataFrame Creation and Export:} A new DataFrame is created from the dictionary of topics. Each row in the DataFrame represents a unique topic, along with its associated concept, ethnicity, topic number, keywords and sentence count. The resulting DataFrame is then exported to a file (extract shown in Table \ref{table:topicsperconceptcounts}) for further analysis and visualisation.

\begin{algorithm}[!h]
\caption{Textual data analysis and topic modeling} \label{alg1}
\begin{algorithmic}[1]
    \State Import data manipulation, text processing, and topic modeling libraries
    \State Download NLTK resources including stopwords, tokenizers, and POS taggers
    \State Load the preprocessed dataset from a csv file
    \Procedure{CleanText}{text}
        \State Remove non-ASCII characters from text
        \State Tokenize text into words
        \State Remove alphanumeric words, keeping alphabetic and numeric
        \State Convert words to lowercase
        \State Remove stopwords and tokens less than two characters
        \State Perform POS tagging, keeping nouns and verbs
        \State \textbf{return} cleaned text
    \EndProcedure
    \State Apply \Call{CleanText}{} to `Sentence' column
    \Procedure{ExtractTopicsLDA}{data, concept, ethnicity}
        \State Subset data based on concept and ethnicity
        \If{subset is empty}
            \State \textbf{return} notification of no data
        \EndIf
        \State Vectorize sentences using TF-IDF
        \State Normalize TF-IDF vectors
        \State Fit LDA model to normalized TF-IDF matrix
        \State Extract top words for each topic
        \State \textbf{return} topics
    \EndProcedure
    \State Remove duplicate topics from list
    \State Create a dictionary for unique topics per concept and ethnicity
    \For{each unique concept and ethnicity combination}
        \State Extract topics using \Call{ExtractTopicsLDA}{}
        \State Remove duplicates and store in dictionary
        \State Count the number of sentences per topic
    \EndFor
    \State Convert topics dictionary to list for DataFrame
    \State Create DataFrame from list
    \State Export DataFrame to a csv file for analysis or presentation
\end{algorithmic}
\end{algorithm}
\vspace{-0.5cm}
\subsubsection{Interactive topic analysis and semantic network visualisation.} This section introduces computational methods designed to: extract and display thematic topics from a dataset, and to visualise the semantic relationships among the keywords of each topic. This dual approach is described as follows. \\\textit{Visualise Topic Relationships}: Complementing the thematic extraction, network diagrams are generated to visualise the semantic relationships among the keywords of each topic. These diagrams employ cosine similarity to map out the connections between words, where nodes signify individual words, and edges reflect the degree of semantic similarity. Such visualisations provide a graphical representation of how topics are compositionally structured, illustrating the semantic proximity and interconnectedness of topic keywords. \\\textit{Interactivity and User Engagement}: Users can specify the concept, ethnicity, number of topics, and similarity threshold, among other parameters, thereby facilitating a highly customised analysis that can adapt to varied research needs.
Algorithm (\ref{NetworkDiagrams}) illustrates the process for constructing topic analysis and network visualisation. The \texttt{calculate\_similarity} function computes the cosine similarity between terms using the TF-IDF matrix. It takes the entire dataset or a subset and calculates the pairwise cosine similarity between all terms. The \texttt{create\_network\_diagram} function constructs the network visualisation. It extracts the top words for each topic and calculates the cosine similarity between their TF-IDF vectors. A new graph (\texttt{G}) is initialised, and for each pair of top words, an edge is added between them if their cosine similarity exceeds a specified threshold. The layout for the network visualisation is calculated, and the graph is drawn using \texttt{matplotlib} with nodes, edges, and labels. The analysis and visualisation are executed by calling the \texttt{extract\_topics\_lda} function to obtain the topics and their top words. For each topic, the \texttt{create\_network\_diagram} function is called with the topic's top words, and the resulting network diagram is visualised.

\begin{algorithm}[h]
\caption{Construct network diagrams from topics} \label{NetworkDiagrams}
\begin{algorithmic}[1]
\Procedure{CalculateSimilarity}{$TFIDFMatrix$}
    \State $SimilarityMatrix \gets$ \Call{CosineSimilarity}{$TFIDFMatrix$}
    \State \Return $SimilarityMatrix$
\EndProcedure

\Procedure{CreateNetworkDiagram}{$Topics$, $SimilarityMatrix$, $Threshold$}
    \For{each $Topic$ in $Topics$}
        \State $TopWords \gets$ \Call{ExtractTopWords}{$Topic$}
        \State $G \gets$ new Graph()
        \For{each pair $(word1, word2)$ in $TopWords$}
            \State $similarity \gets SimilarityMatrix[word1, word2]$
            \If{$similarity > Threshold$}
                \State Add edge $(word1, word2)$ to $G$ with weight $similarity$
            \EndIf
        \EndFor
        \State $Layout \gets$ \Call{CalculateLayout}{$G$}
        \State \Call{DrawGraph}{$G$, $Layout$}
    \EndFor
\EndProcedure

\Procedure{ExtractTopicsAndVisualize}{$DataFrame$, $Concept$, $Ethnicity$, $NumTopics$}
    \State $Topics, TFIDFMatrix \gets$ \Call{ExtractTopicsLDA}{$DataFrame$, $Concept$, $Ethnicity$, $NumTopics$}
    \State $SimilarityMatrix \gets$ \Call{CalculateSimilarity}{$TFIDFMatrix$}
    \State \Call{CreateNetworkDiagram}{$Topics$, $SimilarityMatrix$, $Threshold$}
\EndProcedure
\end{algorithmic}
\end{algorithm}

\vspace{-1cm}
\section{Results}
\subsection{Analysis of concepts across each ethnic group}
Table \ref{table:topicsperconceptcounts} provides an extract of the non-sensitive outputs when running the proposed offline methods on the pre-processed dataset. The table organises LDA analysis results into four columns: the first column details the SIRch concept each topic pertains to, the second specifies the ethnic group associated, the third column lists the topics identified by LDA, and the fourth shows the number of sentences per topic for each ethnic-concept group, providing a thorough overview of topic distribution and relevance within the LDA results. Table \ref{tab:topicbyethnic} illustrates the distribution of five distinct topics across all ethnic groups. \\
The complete version of Table \ref{table:topicsperconceptcounts} is provided in the project's Github repository\footnote{https://github.com/gcosma/I-SIRchpapers/}. The topic analysis results table (i.e. the full version of Table \ref{table:topicsperconceptcounts}) was input into the `Claude 3 Opus' online LLM for providing meaningful analysis and fast processing of non-sensitive data.  The results were human checked and no mistakes were found. Due to space limitations, the discussion that follows will focus on the Asian, Black, and White British ethnic groups.
\vspace{-0.5cm}
\begin{table}[H]
\centering
\caption{Topics in Care Planning (CP) across ethnic groups with the number of sentences per topic (S. Count). WB: White British.}
\begin{tabular}{lllp{2.8in}r}
\hline
\textbf{Conc.}&\textbf{Ethn.} & \textbf{Topic} & \textbf{Keywords} & \textbf{S. Count} \\ \hline
CP&WB & Topic 1 & body, plans, occasions, period, mass, index, inclusion, inform, bmi, mothers & 3 \\
CP& WB & Topic 2 & baby, delivery, mother, birth, benefits, wishes, place, explore, plan, care & 4 \\
CP&WB & Topic 3 & change, gaps, situation, centile, team, weight, timing, labour, documentation, assessments & 3 \\
CP&WB & Topic 4 & pathway, opportunity, risk, pregnancy, staff, management, plan, care, discuss, meant & 8 \\
CP&WB & Topic 5 & monitoring, scenarios, clinicians, guidance, pathways, bp, evidence, triage, addition, advice & 2 \\
CP&Asian & Topic 1 & concerns, ctg, waters, cardiotocograph, times, reasons, possibility, plan, care, delays & 1 \\
CP&Asian & Topic 2 & concerns, ctg, waters, cardiotocograph, days, mother, ongoing, oversight, plan, care & 1 \\
CP& Asian & Topic 3 & ctg, waters, concerns, cardiotocograph, pathway, days, iol, possibility, care, consideration & 3 \\
CP&Black & Topic 1 & appointment, system, mother, treatment, place, plan, safety, care, antihistamines, assessment & 1 \\
CP&Black & Topic 2 & mother, risk, treatment, meant, tolerance, plan, care, antihistamines, assessment, test & 1 \\
CP& Black & Topic 3 & triggers, mother, anaphylaxis, risk, treatment, symptoms, delay, plan, care, antihistamines & 1 \\
CP& Black & Topic 4 & clinician, accumulation, mother, risk, treatment, risks, plan, care, lead, antihistamines & 1 \\
CP&Black & Topic 5 & mother, risk, treatment, meant, management, plan, delay, care, antihistamines, assessment & 2 \\
\hline
\end{tabular}

\label{table:topicsperconceptcounts}
\end{table}

\begin{figure}[!h]
  \centering
\includegraphics[width=\linewidth,height=\textheight,keepaspectratio]{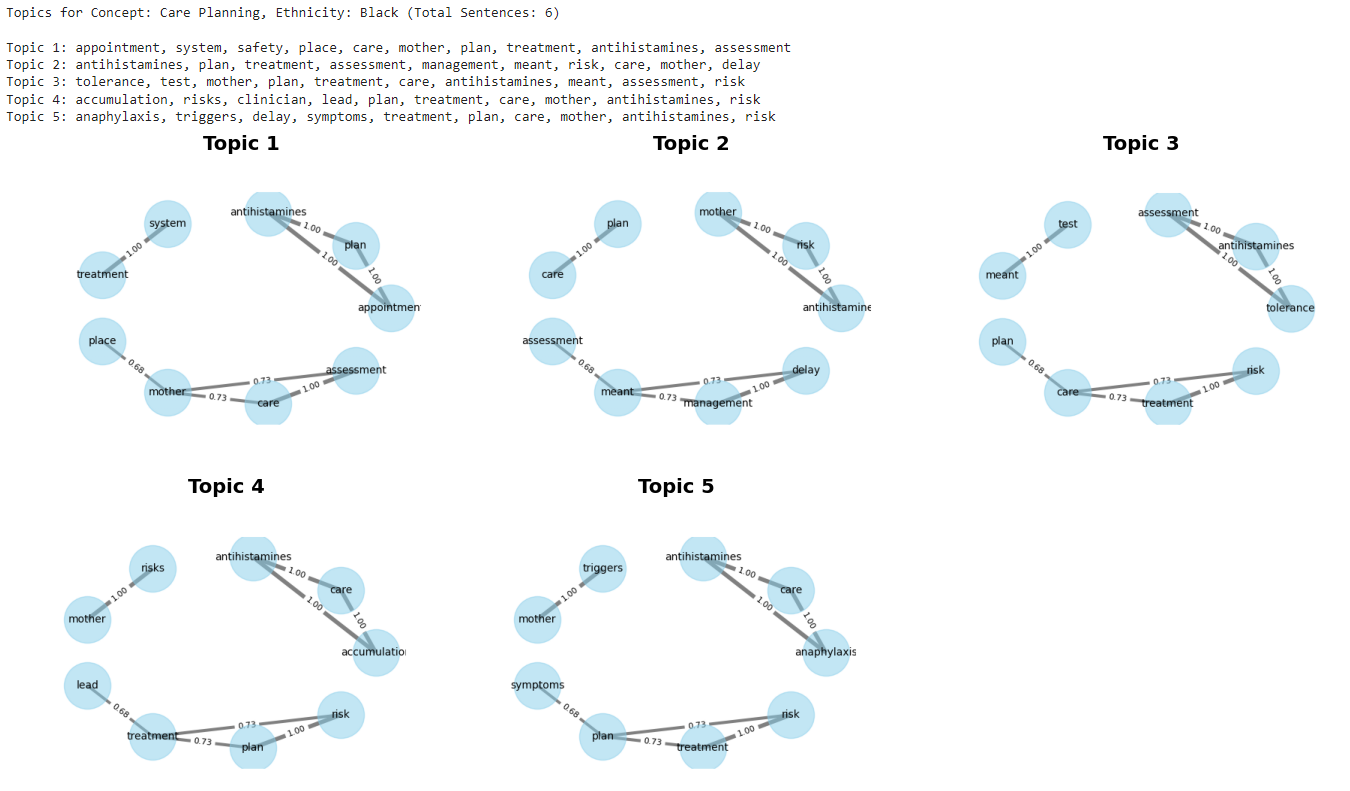}
  \caption{Concept: Care Planning, Ethnicity: Black. Similarity threshold: 0.50} 
  \label{screenshotCare}
\end{figure}

\noindent Fig. \ref{screenshotCare} shows a network diagram for the `Care Planning' concept for the Black ethnic group. Similarity threshold was set to 0.50.
\begin{table}[]
\centering
\caption{Sentence counts by ethnicity and topic. Up to 5 topics per group were generated. Duplicated topics were removed, hence some groups have $<5$ topics.}
\begin{tabular}{lrrrrrr}
\hline
\textbf{Ethnicity} & \textbf{Topic 1} & \textbf{Topic 2} & \textbf{Topic 3} & \textbf{Topic 4} & \textbf{Topic 5} & \textbf{Total sentences} \\ \hline
Asian              & 22               & 24               & 14               & 9                & 10     & 79         \\
Black              & 30               & 21               & 15               & 7                & 4       &   77      \\
Data not received  & 24               & 14               & 10               & 3                & 0        &  51      \\
Mixed Background   & 9                & 2                & 0                & 0                & 0         &   11    \\
Other White        & 21               & 11               & 4                & 1                & 1          & 38     \\
White British      & 132              & 121              & 115              & 179              & 140      & 687       \\ \hline
\end{tabular}
\label{tab:topicbyethnic}
\end{table}
\subsection{Topics around the sentences of the Black ethnic group}
\textbf{Healthcare Processes and Assessments:} The Black ethnic group contributed the most sentences (29) to this topic, which covers various healthcare processes and assessments. This suggests a significant focus on the effectiveness and efficiency of these processes. Concepts such as ``Assessment, investigation, testing, screening'', ``Care Planning'', and ``Risk assessment'' indicate a focus on the quality and comprehensiveness of patient assessments and care planning. Keywords like ``pathway'', ``assessment'', ``care'', ``plan'', and ``review'' relate to the various stages and components of healthcare processes.\\
 \textbf{Patient Care and Management:} The group contributed 24 sentences to this topic, highlighting their concerns regarding patient care and management. This reflects an emphasis on ensuring appropriate and effective care for patients. Concepts such as ``Escalation/referral factor'', ``Psychological characteristics'', and ``Teamworking'' suggest a focus on the various aspects of patient care, including escalation protocols, mental health, and collaboration among healthcare professionals. Keywords like ``care'', ``plan'', ``mother'', ``team'', and ``treatment'' relate to the provision of care and the involvement of different stakeholders in patient management.\\
\textbf{Healthcare Technology and Equipment:} The Black ethnic group contributed 21 sentences related to healthcare technology and equipment. This indicates a moderate level of concern regarding the availability, functionality, and proper use of technology and equipment in healthcare settings. Concepts such as ``Interpretation'', ``Escalation/referral factor'', and ``Teamworking'' suggest a focus on the use of technology for tasks like CTG interpretation, as well as the role of technology in facilitating communication and collaboration among healthcare professionals. Keywords including ``ctg'', ``monitoring'', ``equipment'', and ``resuscitation'' relate to specific technologies and equipment used in patient care.\\
\textbf{COVID-19 Impact on Healthcare:} The group contributed 9 sentences related to the impact of COVID-19 on healthcare. While this topic has fewer sentences compared to others, it still indicates a notable concern regarding the challenges and adaptations required in healthcare settings due to the pandemic. Concepts such as ``COVID'', ``Communication factor'', and ``Teamworking'' suggest a focus on how the pandemic has affected communication, collaboration, and overall functioning of healthcare teams. Keywords like ``resuscitation'', ``staff'', ``care'', ``ppe'', and ``environment'' relate to the specific challenges and changes brought about by COVID-19 in healthcare settings.\\
\textbf{Multidisciplinary Care and Communication:} The group contributed the least number of sentences (6) to this topic, which focuses on multidisciplinary care and communication. Although less prominent than other topics, it still indicates an awareness of the importance of collaboration and effective communication among different healthcare professionals. Concepts like ``Care Planning'' and ``Communication factor'' suggest a focus on the coordination of care and the role of communication in ensuring effective patient management. Keywords such as ``assessment'', ``input'', ``referrals'', ``specialist'', and ``meant'' relate to the involvement of various healthcare professionals in patient care and the communication processes that enable this collaboration.

\subsection{Topics around the sentences of the Asian ethnic group}
\textbf{Emergency Processes:} The Asian ethnic group contributed 22 sentences around emergency processes, indicating a focus in the responsiveness and management of emergency situations. This suggests a concern for how emergencies are handled. Concepts like ``Acuity'', ``Escalation/referral factor'', and ``Situation awareness'' suggest a focus on emergency situations. Keywords such as ``ctg'', ``decisions'', ``concerns'', ``delays'', and ``escalation'' relate to the management of urgent cases.\\
\textbf{Clinical Operations and Equipment:} With 24 sentences, there is a noticeable focus on clinical operations and the use of medical equipment. This number, closely following their contributions to emergency processes, reflects an emphasis on the quality and efficiency of clinical care. Concepts including ``Assessment, investigation, testing, screening'', ``Monitoring'', and ``Technologies and Tools-issues'' indicate a focus on clinical procedures and equipment. Keywords like ``ctg'', ``observation'', ``trace'', ``auscultation'', ``phone'', and ``line'' are associated with medical equipment and clinical tasks.\\
\textbf{Patient Transfer Logistics:} Contributing 14 sentences to this topic, the Asian ethnic group shows a moderate level of concern regarding the logistics involved in patient transfer. This interest points to an understanding of the importance of safe and efficient patient movement within healthcare facilities. The concepts ``Communication factor'', ``Documentation'', and ``Teamworking'' suggest a focus on the coordination and logistics involved in patient care and transfer. Keywords such as ``samples'', ``results'', ``discrepancies'', ``handovers'', and ``format'' relate to the transfer of patient information and the movement of patients within the healthcare system.\\
\textbf{Childbirth Options and Procedures:} With only 9 sentences, there is a lower representation of concerns related to childbirth options and procedures. This might indicate a lesser focus on or fewer experiences reported regarding childbirth compared to other topics. The presence of concepts like ``Care Planning'' and ``Communication factor'' in this topic suggests a focus on the decision-making and communication processes involved in childbirth. Keywords including ``waters'', ``leak'', ``condition'', and ``labour'' are specifically related to childbirth and pregnancy.\\
\textbf{Systemic Healthcare Issues:} The group contributes 10 sentences on systemic healthcare issues, suggesting a moderate concern for overarching healthcare system structures and efficiencies. This indicates an awareness of and concern for the broader healthcare environment. Concepts such as ``Language barrier'' and ``Teamworking'' in this topic indicate a focus on broader, systemic issues in healthcare. Keywords like ``interpreter'', ``trauma'', ``intrapartum'', ``postnatal'', and ``interpreting'' suggest challenges related to communication and support for patients throughout the healthcare system.
\subsection{Topics around the sentences of the White British ethnic group}
\textbf{Clinical Assessments and Monitoring:} The White British ethnic group contributed the most sentences (123) to this topic, which covers clinical assessments, investigations, and monitoring. This suggests a significant focus on the quality and effectiveness of these processes. Concepts such as ``Assessment, investigation, testing, screening'', ``Documentation'', and ``Monitoring'' indicate a strong emphasis on the thoroughness and accuracy of clinical assessments and documentation. Keywords like ``clinicians'', ``risk'', ``observations'', ``examination'', and ``history'' relate to the various components and considerations involved in clinical assessments and monitoring.\\
\textbf{Communication and Interpersonal Factors:} The group contributed 103 sentences to this topic, highlighting the importance of communication and interpersonal factors in healthcare. This reflects a focus on ensuring effective communication among healthcare professionals, patients, and their families. Concepts such as ``Communication factor'', ``Teamworking'', and ``Psychological characteristics'' suggest a recognition of the role of communication, collaboration, and emotional support in providing quality care. Keywords like ``birth'', ``information'', ``family'', ``mother'', and ``team'' relate to the various stakeholders and contexts in which communication and interpersonal factors are crucial.\\
\textbf{COVID-19 Impact on Healthcare:} The White British ethnic group contributed 71 sentences related to the impact of COVID-19 on healthcare. This indicates a significant concern regarding the challenges and adaptations required in healthcare settings due to the pandemic. The concept ``COVID'' is directly related to this topic, suggesting a focus on the specific effects of the pandemic on healthcare processes and outcomes. Keywords such as ``staff'', ``care'', ``maternity'', ``equipment'', and ``impact'' relate to the various aspects of healthcare that have been affected by COVID-19.\\
\textbf{Guidance and Protocol Adherence:} The group contributed 65 sentences related to guidance and protocol adherence. This suggests a notable focus on ensuring that healthcare professionals follow appropriate guidelines and protocols in their practice. Concepts such as ``Local guidance'', ``National guidance'', and ``National and local guidance'' directly relate to the use of established guidelines and protocols in healthcare. Keywords like ``guidance'', ``line'', ``trust'', ``escalation'', and ``management'' relate to the specific aspects of healthcare practice that are guided by established protocols and guidelines.\\
\textbf{Obstetric Care and Decision-Making:} A total of 57 sentences related to obstetric care and decision-making. This indicates a significant focus on the specific challenges and considerations involved in providing quality care in obstetric settings. Concepts such as ``Obstetric review'', ``Decision error'', and ``Escalation/referral factor'' suggest a focus on the critical decision-making processes and potential pitfalls in obstetric care. Keywords like ``birth'', ``pregnancy'', ``mother'', ``baby'', and ``delivery'' relate to the specific context of obstetric care and the key stakeholders involved.

\subsection{Risks in Topic Modelling with Unbalanced Ethnic Representation}
In real-world healthcare data, the under-representation of certain groups often leads to unbalanced datasets that can significantly impact research outcomes. Recognising and addressing these risks is crucial for conducting rigorous and responsible research.\\
\textbf{Bias amplification:} When a dataset predominantly contains sentences from one ethnic group, it may overrepresent the perspectives and experiences of that group, potentially perpetuating stereotypes or inaccuracies. \textit{Mitigation:} Enrich the dataset through targeted recruitment from under-represented groups and employ statistical techniques to adjust for imbalances, ensuring ethical collection of real, representative data.\\
\textbf{Misinterpretation of topics/themes:} Data skew may lead to thematic outcomes that reflect the overrepresented group's concerns, possibly mischaracterising crucial topics relevant to other groups. \textit{Mitigation:} Involve community representatives and domain experts in the analysis to ensure diverse perspectives and to validate topics.\\
\textbf{Ethical concerns:} Analysing unbalanced data without addressing its limitations may inadvertently marginalise already under-represented voices, compromising the ethical imperative for equitable representation. \textit{Mitigation:} Prioritise ethical oversight and community engagement, acknowledging data limitations transparently and seeking consent from under-represented groups throughout the research process.\\
\textbf{Loss of valuable insights:} A more balanced dataset might reveal unique experiences and challenges of under-represented groups, offering new understandings or highlighting areas of concern not visible through the majority group's lens. \textit{Mitigation:} Use existing datasets and conduct supplementary qualitative research to directly gather insights from under-represented communities, employing adaptive sampling methods for qualitative depth.

\section{Conclusion}
\vspace{-0.04cm}
This study illustrates the effectiveness of utilising topic modelling and other NLP techniques to analyse maternity incident investigation reports. It introduces an analytical framework to highlight disparities in care among ethnic groups, pointing out the necessity for focused improvement efforts. The reports underwent data preprocessing, annotation using the SIRch taxonomy, and topic modelling using LDA to identify unique topics based on different combinations of concepts and ethnic groups. To ensure the protection of sensitive data whilst enabling advanced analysis, a combination of offline and online methods was utilised, with offline processing used for sensitive data and online processing for non-sensitive data (i.e. keywords of topics as shown in the Table \ref{table:topicsperconceptcounts} extract) using the `Claude 3 Opus' language model. Interactive topic analysis and semantic network visualisation were employed to extract and display thematic topics and visualise semantic relationships among keywords. \\The research acknowledges potential drawbacks of analysing unbalanced datasets, such as the risk of bias amplification, misinterpretation of themes/topics, ethical issues, and the possible loss of important insights. The findings reveal systemic issues within maternity services, setting the stage for strategies informed by data to prevent incidents and enhance patient care. Involving stakeholders and experts ensures that the insights from the analysis are relevant and contribute to targeted measures to improve healthcare equity and outcomes. A notable limitation of the proposed topic modelling tool is the absence of Patient and Public Involvement (PPI) in evaluating its outputs. PPI plays a vital role, ensuring that the derived solutions are aligned with patient needs and experiences, which could significantly refine the topic summaries and their applicability in real-world scenarios. Incorporating input from patients and the public would likely enhance the model's transparency and trust, making it more reliable. Future enhancements will focus on incorporating PPI feedback to improve the framework's efficiency in topic modelling, ultimately contributing to better patient safety and care quality.

\begin{credits}
\subsubsection{\ackname} This report is independent research funded by NHSX and The Health Foundation and it is managed by the National Institute for Health Research (AI\_HI200006). The views expressed in this publication are those of the author(s) and not necessarily those of the NHSX, The Health Foundation, National Institute for Health Research, or the Department of Health and Social Care.

\subsubsection{\discintname}
The authors have no competing interests to declare that are
relevant to the content of this article. 
\end{credits}

\bibliographystyle{splncs04}
\bibliography{references}

\end{document}